\documentclass[letterpaper, 10 pt, conference]{ieeeconf}  

\IEEEoverridecommandlockouts                              

\usepackage{amsmath,epsfig}
\usepackage{multirow}

\usepackage[natbib, bibencoding=utf8, citestyle=numeric, bibstyle=ieee, maxbibnames=999, maxcitenames=2, mincitenames=1, sortcites]{biblatex}

\usepackage{tabularx}
\usepackage{tabulary}
\usepackage{makecell}

\usepackage{multirow}
\usepackage{subcaption}
\usepackage{booktabs}
\usepackage{threeparttable}

\bibliography{refs.bib}

\newcommand{\eg}{e.\,g. }
\newcommand{\ie}{i.\,e. }

\usepackage[capitalise, nameinlink, noabbrev]{cleveref}
\usepackage{tabularx}
\usepackage{makecell}
\title{\LARGE \bf
Depression Diagnosis and Forecast based on Mobile Phone Sensor Data
}

\author{Xiangheng He$^{1,2*}$, Andreas Triantafyllopoulos$^{1}$, Alexander Kathan$^{1}$, Manuel Milling$^{1}$, Tianhao Yan$^{1}$\\Srividya Tirunellai Rajamani$^{1}$, Ludwig Küster$^{3}$, Mathias Harrer$^{3,4}$, Elena Heber$^{3}$, Inga Grossmann$^{3}$,\\ David D. Ebert$^{3,4}$, Björn W.\ Schuller$^{1,2}$
\thanks{$^{1}$X.\,H., A.\,T., A.\,K., M.\,M., T.\,Y., S.\,T.\,R., and B.\,S. are with the Chair of Embedded Intelligence for Health Care \& Wellbeing, University of Augsburg, Germany}%
\thanks{$^{2}$B.\,S. and X.\,H. are also with GLAM -- the Group on Language, Audio,
\& Music, Imperial College London, London, UK}
\thanks{$^{3}$L.\,K., M.\,H., E.\,H., I.\,G. and D.\,E. are with GET.ON Institut für Online Gesundheitstrainings GmbH/HelloBetter, Hamburg, Germany}
\thanks{$^{4}$M.\,H., and D.\,E. are also with the Chair of Psychology \& Digital Mental Health Care, Technical University Munich, Germany}
\thanks{$^{*}$ Corresponding author:
        {\tt\small x.he20@imperial.ac.uk}}%
\thanks{© 2022 IEEE.  Personal use of this material is permitted.  Permission from IEEE must be obtained for all other uses, in any current or future media, including reprinting/republishing this material for advertising or promotional purposes, creating new collective works, for resale or redistribution to servers or lists, or reuse of any copyrighted component of this work in other works.}
}

\begin{document}\sloppy

\maketitle
\thispagestyle{empty}
\pagestyle{empty}

\begin{abstract}
Previous studies have shown the correlation between sensor data collected from mobile phones and human depression states. Compared to the traditional self-assessment questionnaires, the passive data collected from mobile phones is easier to access and less time-consuming. In particular, passive mobile phone data can be collected on a flexible time interval, thus detecting moment-by-moment psychological changes and helping achieve earlier interventions. Moreover, while previous studies mainly focused on \emph{depression diagnosis} using mobile phone data, \emph{depression forecasting} has not received sufficient attention. In this work, we extract four types of passive features from mobile phone data, including phone call, phone usage, user activity, and GPS features. We implement a long short-term memory (LSTM) network in a subject-independent 10-fold cross-validation setup
to model both a diagnostic and a forecasting tasks. Experimental results show that the forecasting task achieves comparable results with the diagnostic task, which indicates the possibility of forecasting depression from mobile phone sensor data. Our model achieves an accuracy of 77.0\,\% for major depression forecasting (binary), an accuracy of 53.7\,\% for depression severity forecasting (5 classes), and a best RMSE score of 4.094 (PHQ-9, range from 0 to 27). 
\end{abstract}

\section{Introduction}

\begin{table*}[]
\caption{A total of 19 features extracted and their descriptions.}
\begin{tabular}{lll}
\hline
\textbf{Data type}                            & \textbf{Extracted feature}                       & \multicolumn{1}{l}{\textbf{Description}}                                                                                                                                                                                                                                                \\ \hline
{}                                     & {total calling frequency}            & the number of times that a participant answers and makes phone calls during a day.                                                                                                                                                                                               \\ \cline{2-3} 
{}                                     & {total calling duration}             & \makecell[l]{the total time in minutes that a participant spends each day answering and \\ making phone calls.}                                                                                                                                                                                   \\ \cline{2-3} 
{}                                     & {non-working time calling frequency} &\makecell[l]{the number of times that a participant answers and makes phone calls at times other\\ than 8 am to 6 pm during a day.}                                                                                                                                                              \\ \cline{2-3} 
{}                                     & {non-working time calling duration}  & \makecell[l]{the total time that a participant answers and makes phone calls at times other than\\ 8am to 6pm during the day.}                                                                                                                                                                   \\ \cline{2-3} 
{}                                     & {number of missed calls}             & the number of calls that are marked as missed during the day.                                                                                                                                                                                                                    \\ \cline{2-3} 
{}                                     & {number of contacts}                 & the number of contact a participant answers and makes phone calls during the day.                                                                                                                                                                                                \\ \cline{2-3} 
{}                                     & {calling entropy}                    & the variability of calling durations a participant spends in contacts during the day. 
\\ \cline{2-3} 
\multirow{-8}{*}{{\textbf{Phone call data}}}    & {normalised calling entropy}         & calling entropy divided by the logarithm of the number of contacts during the day.                                                                                                                                                                                               \\ \hline
{}                                     & {phone usage frequency~\citep{moshe2021predicting}}              & the number of times that a participant interacts with their phone during a day.                                                                                                                                                                                                  \\ \cline{2-3} 
\multirow{-2}{*}{{\textbf{Phone usage data}}}   & {phone usage duration~\citep{moshe2021predicting}}               & \makecell[l]{the total time in seconds that participants spend each day interacting with their mobile\\ phones.}                                                                                                                                                                                 \\ \hline
{}                                     & {lock screen duration}               & the total time in seconds that participants lock their mobile phones during the day.                                                                                                                                                                                             \\ \cline{2-3} 
{}                                     & {number of used apps}                & the number of applications that a participant uses during the day.                                                                                                                                                                                                               \\ \cline{2-3} 
{}                                     & {number of midnight used apps}       & the number of applications that a participant uses between 0am to 5am during the day.                                                                                                                                                                                            \\ \cline{2-3} 
\multirow{-4}{*}{{\textbf{User activity data}}} & {sleep time}                         & \makecell[l]{sleep time is considered to be between the last time an app was used in the previous\\ day (or in the same day before 2\,am if available) and the first time an app was used\\ after 5\,am}                                                                                                                                                                                                                                                                           \\ \hline
{}                                     & {location variance~\citep{moshe2021predicting}}                  & \makecell[l]{the logarithm of the sum of the statistical variances in the latitude and the longitude of\\ all GPS coordinates in the day.}                                                                                                                                                       \\ \cline{2-3} 
{}                                     & {location entropy~\citep{moshe2021predicting}}                   & the variability of the time that participants spend in significant places in the day.                                                                                                                                                                         \\ \cline{2-3} 
{}                                     & {normalised location entropy~\citep{moshe2021predicting}}        & the location entropy divided by the logarithm of the number of significant places.                                                                                                                                                                                               \\ \cline{2-3} 
{}                                     & {time at home~\citep{moshe2021predicting}}                       & \begin{tabular}[c]{@{}l@{}} \makecell[l]{Home is defined as the most frequent significant place where a participant spent\\ the most time between 0\,am to 6\,am.}\\ \makecell[l]{Time at home is defined as the percentage of time a participant spent at home\\ relative to other significant places.}\end{tabular} \\ \cline{2-3} 
\multirow{-5}{*}{{\textbf{GPS data}}}           & {total distance~\citep{moshe2021predicting}}                     & the total distance covered by a participant during the day.                                                                                                                                                                                                          \\ \hline
\end{tabular}
\label{Extracted_features}
\end{table*}

Depression, as a common mental health disorder, is typically characterised by low mood, overthinking, feelings of hopelessness, and decreased motivation. In extreme cases, people experiencing severe depression may have suicidal thoughts. Depression affects not only individual patients and their families, but also their social circle and overall economic development~\citep{auerbach2016mental}. In Germany, depression is the leading cause of the inability to work or early retirement and is the trigger for about half of all suicides each year. While most people with depression are treated in primary care settings, more than 50\,\% of people are not identified or effectively treated~\citep{moshe2021predicting}. 

The long-lasting primary method of clinical depression diagnosis relies on the self-assessment questionnaires, such as the Patient Health Questionnaire (PHQ)-2 and PHQ-9. These questionnaires have shown a strong correlation with actual human health. However, collecting them is usually time-consuming and has fixed time intervals, which can hardly detect the moment-by-moment psychological changes and achieve timely interventions.

Recently, the rise of wearable devices and mobile phones has made sensor data more readily available. Previous studies have explored the possibility of using sensor data to diagnose human mental health states and have shown the effectiveness~\cite{han2021deep, qian2021artificial}. \citet{rohani2018correlations} provided a systematic survey for the correlations between sensor data and depressive mood symptoms. Compared to the self-assessment questionnaires, the passive data collection does not require an interaction with the device and can be collected at a more flexible time interval, which means that it can reflect immediate changes in psychological state, potentially enabling early diagnosis, prediction of disease progression, and timely adjustment of treatment plans.

However, previous studies mainly focused on \emph{depression diagnosis} based on mobile phone data, \ie, the prediction of depression state and/or severity for a given time-period (\eg, on a daily, weekly, or bi-weekly basis) given concurrent features. In contrast, the forecasting of depression progression,  \ie, the prediction of state/severity on a given time-period given features further in its past, has not received sufficient attention. \citet{saeb2015mobile} has shown the effectiveness of using features extracted from mobile phone GPS and usage of sensors to diagnose if participants have depressive symptoms (PHQ-9$\geq$5). \citet{masud2020unobtrusive} extracted 12 features from GPS and acceleration data and classified participants' weekly PHQ-9 into three groups based on that week's features. \citet{lu2018joint} used the GPS, activity, sleep, and heart rate data collected from mobile phones and wearable devices to distinguish the participants with depression and diagnosed their clinical severity. They also only used the features from the same week to make the weekly diagnosis.

Different from previous work, we design two tasks for both diagnosis and forecasting: the first task is to diagnose the current week's PHQ-9 score according to data from the same week, while the second task is to forecast the PHQ-9 score at the end of next week based on data from the current week. We treat the diagnosis and forecasting of PHQ-9 as a regression problem and implement an LSTM model combined with a subject-independent 10-fold cross-validation. 
We use a portion of passive data from a newly collected dataset called MAIKI,
which includes phone call, phone usage, user activity, and GPS data. We choose root-mean-square error (RMSE)
as the evaluation metric and use two methods categorising PHQ-9 scores into different subgroups. We distinguish the participants with major depression (PHQ-9$\geq$10) from those without. Additionally, we report a 5-class depression severity. Results show that the forecasting task achieves comparable results with the diagnostic task, which indicates the possibility of forecasting depression from mobile phone data. In order to compare different algorithm options and parameter settings, we also compare three different clustering methods to identify significant places (GPS coordinates that need to be considered the same place and meet certain conditions) from the GPS data. 

The rest of the paper is organised as follows. \cref{sec:maikidataset} introduces the newly collected MAIKI dataset and our feature extraction methods. \cref{sec:experimentalsetup} describes our task design, experimental setting, and evaluation approaches. \cref{sec:results} outlines the obtained results for diagnostic and forecasting tasks. \cref{sec:conclusion} concludes the paper with a brief discussion.

\section{Dataset and feature extraction}
\label{sec:maikidataset}
\subsection{MAIKI dataset}
The MAIKI dataset is collected from the ``Mobile daily living therapy assistant with interaction-focused artificial intelligence for depression'' (MAIKI) project. A total of 48 people 
participated in this project and carried mobile phones with a sensor data acquisition app for 8 weeks. The study procedures were approved by the ethics committee of the Friedrich-Alexander-University Erlangen-Nuremberg (385\_20B). The dataset has both active data from self-assessment questionnaires and passive data from mobile phone sensors. The active questionnaire data includes the weekly PHQ-9 and other questionnaires data such as Generalised Anxiety Disorder (GAD-7) and Perceived Stress Scale (PSS-4).
The passive data includes phone call, phone usage, user activity, GPS, battery, phone text, ringtone setting, and step count data, which were collected during each day.

In this work, we focus only on using (parts of) the passive data, namely phone call, phone usage, user activity, and GPS data 
to diagnose and forecast the weekly PHQ-9 scores. \cref{Extracted_features} shows an overview over all features per data type. In \cref{subsec:features}, we outline the procedure followed to extract the features of each data type, placing an emphasis on GPS features which follow a more involved process.

\subsection{Feature extraction}
\label{subsec:features}
\subsubsection{Phone call, phone usage, and user activity features}
We extract a total of 8, 2, and 4 features from phone call, phone usage, and user activity data, respectively.
The descriptions of these features can be seen in \cref{Extracted_features}. These features are all extracted at a daily level for each participant.

\subsubsection{GPS features}
The extraction of GPS features is composed of three steps. The first step is to preprocess the raw GPS data. We remove GPS coordinates with positioning accuracy $\textgreater$80th percentile of all participants' GPS accuracy and additionally remove GPS measurements taken at a speed less than 0. 
Since only the GPS coordinates in the stationary state should be used in the following clustering step to identify significant places,
we remove the GPS coordinates in the transition state with a speed of more than 1.4\,m/s. The second step aims to aggregate the GPS coordinates of the same location into a cluster. The cluster that meet certain conditions is considered a significant place. To compare different algorithm options and parameter settings, we implement three clustering algorithms described below.

\textbf{Time-based clustering.} The basic idea of this algorithm is to cluster the GPS coordinates along the time axis and remove the intermediate coordinates between significant places~\cite{kang2005extracting}. This algorithm computes the place clusters incrementally as the next GPS coordinates come in. The algorithm has two parameters: the distance threshold $D_{time}$ is the maximum distance at which the next coordinate is considered to belong to the current place cluster; the time threshold $T_{time}$ is the minimum time duration for which the current cluster is considered as a significant place. When a cluster is a significant place, the algorithm checks whether the cluster should be merged into one of the existing clusters according to the distance between their centroids (the merged distance threshold equals $D_{time}$/3).

\textbf{K-Means clustering.} The typical K-Means clustering algorithm requires a predetermined number of clusters $k$. But in our cases, the number of clusters can vary widely among different participants. Following \citet{saeb2015mobile}, 
we first set $k$ to 1 and increase the cluster number until the distance of the farthest point in each cluster to its cluster centre is less than a threshold $D_{kmeans}$. This threshold determines the maximum radius of a cluster.

\textbf{DBSCAN clustering.} The  Density-Based Spatial Clustering of Applications with Noise (DBSCAN) algorithm has two parameters: the $eps$ is the maximum distance between two coordinates for one to be considered as belonging to the same cluster of the other; the $minSamples$ is the minimum number of data points to form a cluster. This algorithm is generally regarded as particularly suitable for GPS data, as it allows to identify clusters of varying shapes and is robust to outliers~\cite{muller2021depression}. 

The third step is extracting the GPS features. The description of these features can be found in \cref{Extracted_features}. We perform clustering algorithms on all days of data for each participant, then extract GPS features for each participant on each day.

\section{Experimental setup}
\label{sec:experimentalsetup}
We design two tasks for diagnosis and forecasting. The first task is to diagnose the current week's PHQ-9 score according to data from the same week. The second task is to forecast the PHQ-9 score at the end of next week based on data from the current week. For example, the diagnostic task is to predict the PHQ-9 score on day 7 based on the data from day 1 to day 7. In contrast, the forecasting task is to predict the PHQ-9 score on day 14 based on the data from day 1 to day 7. In order to maximise the utilisation of day-level features, we do not use the average of features in a week but give the same weekly PHQ-9 score as the label to the daily data in that week. We treat these two tasks as a regression problem and implement an LSTM model combined with a subject-independent 10-fold cross-validation
to complete these tasks. The model has one LSTM layer, one fully connected layer, and a ReLU activation function. 
The learning rate and the hidden size of LSTM are set to 0.001 and 4, respectively. We choose the mean squared error as the loss function. The model is trained by gradient descent and using the Adam optimiser with $\beta$1 and $\beta$2 set to 0.9 and 0.999.

As for the time-based clustering algorithm for GPS features, since the GPS data from our dataset is recorded every 5 minutes, we set $T_{time}$ to 15\,minutes and $D_{time}$ to 40\,metres~\cite{kang2005extracting}. 
For k-means clustering, we set $D_{kmeans}$ to 500\,metres~\cite{saeb2015mobile}. 
For DBSCAN, we set $eps$ to 30\,metres and $minSamples$ to 3~\cite{muller2021depression}. 
We use Haversine distance as the distance function.

We choose the RMSE as the evaluation metric for regression and utilise two methods categorising PHQ-9 scores into subgroups. We set the cutoff value to 10 to distinguish the participants with major depression (PHQ-9$\geq$10)~\cite{kroenke2001phq} 
from those without and report the 2-class classification accuracy. We evaluate the predicted severity of depression according to \cref{severity} and report the 5-class classification accuracy. 

\begin{table}[t]
\begin{center}
\caption{Scale and 5-class scheme of depression severity.} 
\begin{tabular}{cl}
\hline
\textbf{PHQ-9 Score} & \textbf{Depression Severity} \\ \hline
0--4                  & Minimal depression           \\ 
5--9                  & Mild depression              \\ 
10--14                & Moderate depression          \\ 
15--19                & Moderately severe depression \\ 
20--27                & Severe depression            \\ \hline
\end{tabular}
\label{severity}
\end{center}
\end{table}

\section{Results}
\label{sec:results}




\begin{figure}[t]
    \centering
    \includegraphics[width=.49\textwidth]{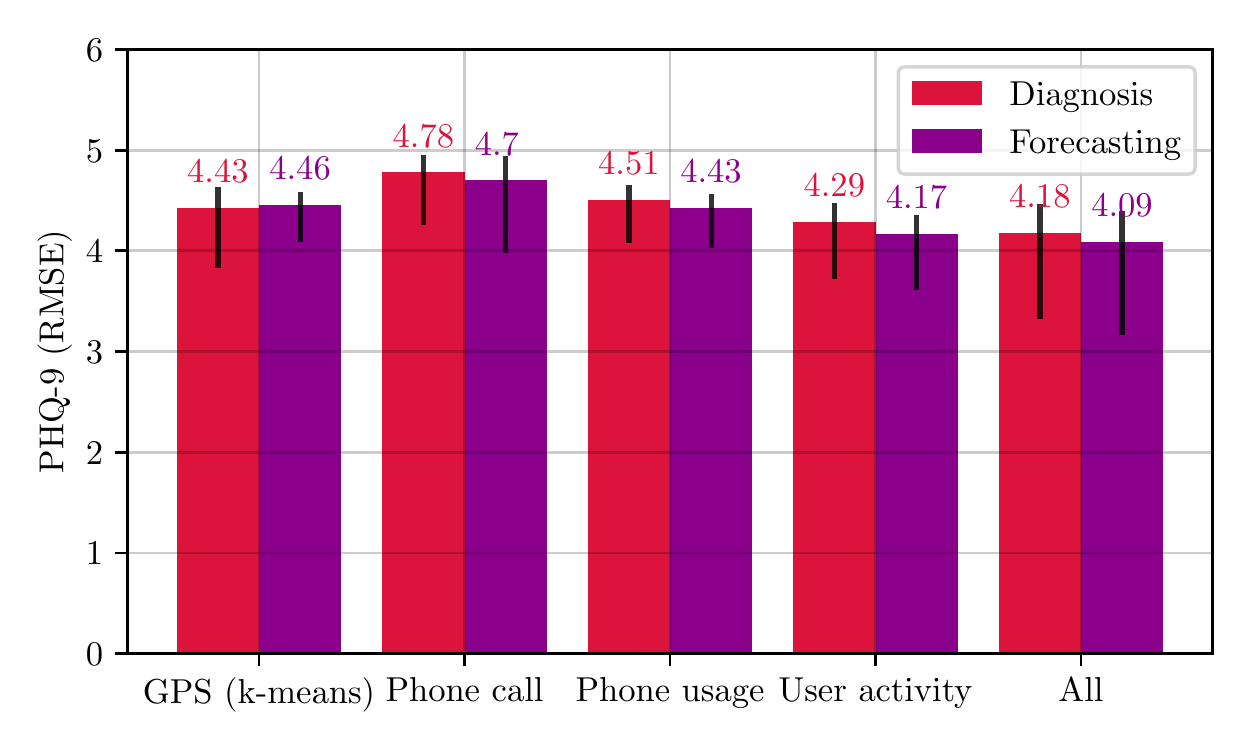}
    \caption{
    Performance comparison for PHQ-9 (range from 0 to 27) diagnosis (left) vs forecasting (right) using different feature sets.
    }
    \label{fig:features}
\end{figure}

\begin{table*}[t]
\centering
\caption{
Results of diagnosing/forecasting the PHQ-9 (range from 0 to 27) at the end of the current/next week based on the data of this week.
Accuracy [\%] reported for major depression (binary) and depression severity (5-class) tasks whereas RMSE is reported for PHQ-9 prediction.
Baseline computed using mean PHQ-9 score of the training set.
Mean performance and the standard deviations (in brackets) are reported over all 10 folds.
}
\label{results}
\begin{tabular}{l|rr|rr|rr}
\toprule
& \multicolumn{2}{c}{\thead{Major Depression (\%\,Acc.)}} & \multicolumn{2}{c}{\thead{Depression Severity (\%\,Acc.)}} & \multicolumn{2}{c}{\thead{PHQ-9 (RMSE)}}  \\
\thead{Model} & \thead{Diagnosis} & \thead{Forecasting} & \thead{Diagnosis} & \thead{Forecasting} & \thead{Diagnosis} & \thead{Forecasting}\\
\midrule
Baseline & 60.6 & 56.7 & 49.7 & 43.0 & 4.858 & 4.915\\ 
K-means & {78.4 (3.5)} & \textbf{77.0 (6.7)} & \textbf{54.5 (7.1)} & \textbf{53.7 (6.4)} & \textbf{4.184 (0.569)} & \textbf{4.094 (0.619)}\\ 
DBSCAN & {74.4 (7.2)} & {71.9 (6.6)} & {52.5 (5.3)} & {47.4 (7.8)} & {4.443 (0.431)} & {4.401 (0.349)}\\ 
Time-based & \textbf{78.9 (4.6)} & {75.7 (5.3)} & {54.5 (4.7)} & {48.5 (7.3)} & {4.203 (0.621)} & {4.556 (0.445)}\\
\bottomrule
\end{tabular}
\end{table*}

\subsection{Results of diagnostic task}
\cref{results} shows the results of diagnosis the current week's PHQ-9 based on data from the same week. We calculate the baseline using the mean value of PHQ-9. Specifically, we assume that all the predictions from the baseline model are the
mean value and then use this assumed mean prediction to calculate the RMSE with the actual labels. The results of all three methods are better than the baseline model. The optimal result is obtained from K-Means, which achieves an accuracy of 78.4\,\% for major depression diagnosis, $54.5\,\%$ for depression severity diagnosis, and a best RMSE score of 4.184. 
The time-based clustering algorithm obtains suboptimal results. It is worth mentioning that, contrary to previous findings that the DBSCAN clustering algorithm may be more suitable for GPS data~\cite{muller2021depression};  
DBSCAN obtained the worst results in our setting.
\subsection{Results of forecasting task}
\cref{results} additionally shows the results of forecasting the PHQ-9 score at the end of next week based on data from the current week. The K-Means algorithm still obtains optimal results, which achieves an accuracy of 77.0\,\% for major depression forecasting and 53.7\,\% for depression severity forecasting. The best RMSE score of 4.094 is marginally lower than that of the diagnostic task, which means that the forecasting task achieves comparable results with the diagnostic task. The results indicate that it is possible to forecast depression based on mobile phone data. 

\subsection{Feature comparison}
Finally, in \cref{fig:features} we show a performance comparison for PHQ-9 diagnosis and forecasting using different features.
We observe that the user activity features obtain the best individual performance for both tasks.
The performance of the phone call features is worse, potentially because this data is more sparse as subjects accept and conduct calls less frequently than using the phone for other purposes.
Combining all information in the all feature sets results in a slight performance boost. 
\section{Conclusion}
\label{sec:conclusion}

In this work, we investigated the potential of using passively collected mobile phone data for depression diagnosis and forecasting.
We have shown that forecasting (predicting PHQ-9 scores, major depression, and depression severity) 1-week ahead of the collected data is possible, with performance close to that of predicting the current week (diagnosis). 
These results showcase the potential of such features of timely diagnosis and change-of-state prediction; both valuable targets for future digital health applications.
These tasks are best modelled using a combination of features, of which user activity are the most important.
In addition, our experiments show that K-Means clustering for generating GPS features fares better than DBSCAN and time-based clustering; a finding which contrasts previous work showing the latter to be better.
This indicates that the performance of such algorithms might be dataset-dependent, and thus a cross-study comparison would be necessary to identify the strengths and weaknesses of each for depression detection.

\section*{Acknowledgments}
\noindent
Data analysed in this publication were collected as part of the MAIKI project, which was funded by the German Federal Ministry of Education and Research (grant No.\ 13GW0254). The responsibility for the content of this publication lies with the authors.


\section{\refname}
\printbibliography[heading=none]

@article{auerbach2016mental,
  title={Mental disorders among college students in the World Health Organization world mental health surveys},
  author={Auerbach, Randy P and Alonso, Jordi and Axinn, William G and Cuijpers, Pim and Ebert, David D and Green, Jennifer G and Hwang, Irving and Kessler, Ronald C and Liu, Howard and Mortier, Philippe and others},
  journal={Psychological medicine},
  volume={46},
  number={14},
  pages={2955--2970},
  year={2016},
  publisher={Cambridge University Press}
}

@article{kroenke2001phq,
  title={The PHQ-9: validity of a brief depression severity measure},
  author={Kroenke, Kurt and Spitzer, Robert L and Williams, Janet BW},
  journal={Journal of general internal medicine},
  volume={16},
  number={9},
  pages={606--613},
  year={2001},
  publisher={Wiley Online Library}
}

@article{muller2021depression,
  title={Depression predictions from GPS-based mobility do not generalize well to large demographically heterogeneous samples},
  author={M{\"u}ller, Sandrine R and Chen, Xi Leslie and Peters, Heinrich and Chaintreau, Augustin and Matz, Sandra C},
  journal={Scientific Reports},
  volume={11},
  number={1},
  pages={1--10},
  year={2021},
  publisher={Nature Publishing Group}
}

@article{kang2005extracting,
  title={Extracting places from traces of locations},
  author={Kang, Jong Hee and Welbourne, William and Stewart, Benjamin and Borriello, Gaetano},
  journal={ACM SIGMOBILE Mobile Computing and Communications Review},
  volume={9},
  number={3},
  pages={58--68},
  year={2005},
  publisher={ACM New York, NY, USA}
}

@article{masud2020unobtrusive,
  title={Unobtrusive monitoring of behavior and movement patterns to detect clinical depression severity level via smartphone},
  author={Masud, Mohammed T and Mamun, Mohammed A and Thapa, K and Lee, DH and Griffiths, Mark D and Yang, S-H},
  journal={Journal of biomedical informatics},
  volume={103},
  pages={103371},
  year={2020},
  publisher={Elsevier}
}

@article{lu2018joint,
  title={Joint modeling of heterogeneous sensing data for depression assessment via multi-task learning},
  author={Lu, Jin and Shang, Chao and Yue, Chaoqun and Morillo, Reynaldo and Ware, Shweta and Kamath, Jayesh and Bamis, Athanasios and Russell, Alexander and Wang, Bing and Bi, Jinbo},
  journal={Proceedings of the ACM on Interactive, Mobile, Wearable and Ubiquitous Technologies},
  volume={2},
  number={1},
  pages={1--21},
  year={2018},
  publisher={ACM New York, NY, USA}
}

@article{rohani2018correlations,
  title={Correlations between objective behavioral features collected from mobile and wearable devices and depressive mood symptoms in patients with affective disorders: systematic review},
  author={Rohani, Darius A and Faurholt-Jepsen, Maria and Kessing, Lars Vedel and Bardram, Jakob E},
  journal={JMIR mHealth and uHealth},
  volume={6},
  number={8},
  pages={e165},
  year={2018},
  publisher={JMIR Publications Inc., Toronto, Canada}
}

@article{moshe2021predicting,
  title={Predicting Symptoms of Depression and Anxiety Using Smartphone and Wearable Data},
  author={Moshe, Isaac and Terhorst, Yannik and Asare, Kennedy Opoku and Sander, Lasse Bosse and Ferreira, Denzil and Baumeister, Harald and Mohr, David C and Pulkki-R{\aa}back, Laura},
  journal={Frontiers in psychiatry},
  volume={12},
  year={2021},
  publisher={Frontiers Media SA}
}

@article{saeb2015mobile,
  title={Mobile phone sensor correlates of depressive symptom severity in daily-life behavior: an exploratory study},
  author={Saeb, Sohrab and Zhang, Mi and Karr, Christopher J and Schueller, Stephen M and Corden, Marya E and Kording, Konrad P and Mohr, David C},
  journal={Journal of medical Internet research},
  volume={17},
  number={7},
  pages={e175},
  year={2015},
  publisher={JMIR Publications Inc., Toronto, Canada}
}

@article{han2021deep,
  title={Deep learning for mobile mental health: challenges and recent advances},
  author={Han, Jing and Zhang, Zixing and Mascolo, Cecilia and Andr{\'e}, Elisabeth and Tao, Jianhua and Zhao, Ziping and Schuller, Bj{\"o}rn W},
  journal={IEEE Signal Processing Magazine},
  volume={38},
  number={6},
  pages={96--105},
  year={2021},
  publisher={IEEE}
}

@article{qian2021artificial,
  title={Artificial intelligence internet of things for the elderly: From assisted living to health-care monitoring},
  author={Qian, Kun and Zhang, Zixing and Yamamoto, Yoshiharu and Schuller, Bjoern W},
  journal={IEEE Signal Processing Magazine},
  volume={38},
  number={4},
  pages={78--88},
  year={2021},
  publisher={IEEE}
}


\end{document}